\documentclass[]{IEEEtran}
\usepackage{graphicx}
\usepackage{amsmath}
\usepackage{array}
\usepackage{times}
\usepackage{epsfig}
\usepackage{amssymb}
\usepackage{epstopdf}
\usepackage{float}
\usepackage{url,multirow}
\usepackage[margin=10pt, textfont=it,labelfont={bf, rm}]{caption}
\usepackage{chngpage}

\ifCLASSINFOpdf
\else
\fi

\begin{document}
\title{Estimating Depth from Monocular Images as Classification Using Deep Fully Convolutional Residual Networks
}
\author{Yuanzhouhan Cao, Zifeng Wu, Chunhua Shen
    		\thanks{Y. Cao, Z. Wu and C. Shen are with  The University of Adelaide,  School of Computer Science, Australia. (email:\{\texttt{yuanzhouhan.cao, zifeng.wu, chunhua.shen\}@adelaide.edu.au.})}
}

\IEEEoverridecommandlockouts
 \IEEEpubid{\begin{minipage}{\textwidth}\ \\ \\ \\[12pt]
Appearing in
IEEE Transactions on Circuits and Systems for Video Technology.
Copyright \copyright 2017 IEEE. 
Personal use of this material is permitted. However, permission to use this material for any other purposes must be obtained from the IEEE by sending an email to \texttt{pubs-permissions@ieee.org}. 
\end{minipage}} 

\maketitle

\begin{abstract}
Depth estimation from single monocular images is a key component in scene understanding. Most existing algorithms formulate depth estimation as a regression problem due to the continuous property of depths. However, the depth value of input data can hardly be regressed exactly to the ground-truth value. In this article, we propose to formulate depth estimation as a pixel-wise classification task. Specifically, we first discretize the continuous ground-truth depths into several bins and label the bins according to their depth ranges. Then we solve the depth estimation problem as classification by training a fully convolutional deep residual network. Compared to estimate the exact depth of a single point, it is easier to estimate its depth range. More importantly, by performing depth classification instead of regression, we can easily obtain the confidence of a depth prediction in the form of probability distribution. With this confidence, we can apply an information gain loss to make use of the predictions that are close to ground-truth during training, as well as fully-connected conditional random fields (CRF) for post-processing to further improve the performance. We test our proposed method on both indoor and outdoor benchmark RGB-D datasets and achieve state-of-the-art performance.

\end{abstract}

\IEEEpeerreviewmaketitle

\tableofcontents
\clearpage

\section{Introduction}

Depth estimation is one of the most fundamental tasks in computer vision. Many other computer vision tasks such as object detection, semantic segmentation, scene understanding, can benefit considerably from accurate estimation of depth information. Most existing methods \cite{Eigen15,LiB15,Wang_2015_CVPR,LiuSLR15} formulate depth estimation as a structured regression task due to the fact of depth values being continuous. These regression models for depth estimation are trained by iteratively minimizing the $L2$ norm between the predicted depths and the ground-truth depths, and aim to output depths as close to the actual depths as possible during evaluation. However, it is difficult to regress the depth value of input data to be exactly the ground-truth value. For human beings, we may find it difficult to tell the exact distance of a specific point in a natural scene, but we can easily give a rough distance range of that point. Motivated by this, we formulate depth estimation as a  pixel-wise classification task by discretizing the continuous depth values into several discrete bins. Instead of training a model to predict the depth value of a point, we train a model to predict the depth range. We show that this simple re-formulation scheme performs surprisingly well.

Another important reason for us to choose classification over regression for depth estimation is that it naturally predicts a confidence in the form of probability distribution over the output space. Different points have different distributions of possible depth values. The depth estimation of some points are easy while others are not. Typical regression models only output the mean values of possible depth values without the variances, (i.e., the confidence of a prediction is missing). Some efforts have been made to obtain this confidence such as the constrained structured regression \cite{deepak16}, or the Monte-Carlo dropout \cite{bayesiansegnet,Gal2016Bayesian}. Compared to  these methods which either require specific constraints or multiple forward passes during evaluation, our proposed approach is simple to implement. 

The obtained probability distribution can be an important cue during both training and post-processing. Although we formulate depth estimation as a classification task by discretization, the depth labels are different from the labels of typical classification tasks such as semantic segmentation. During training, the predicted depth labels that are close to ground-truth and with high confidence  can also be used to update model parameters. This is achieved by an information gain loss. As for the post-processing, we apply the fully-connected conditional random fields (CRF) \cite{phil2011} which have been frequently applied in semantic segmentation \cite{LinSRH15,ChenPKMY14}. With the fully connected CRFs, pixel depth estimation with low confidence can be improved by other pixels that are connected to it.

Traditional depth estimation methods enforce geometric assumptions and rely on hand-crafted features such as SIFT, PHOG, GIST, texton, etc. Recently, computer vision has witnessed a series of breakthrough results introduced by deep convolutional neural networks (CNN) \cite{NIPS2012_4824, Simonyan14c}. The success of deep networks can be partially attributed to the rich features captured by the stacked layers. Recent evidence has shown that depth estimation benefits largely from increased number of layers \cite{EigenPF14, Eigen15, LiuSLR15}. However,  stacking more layers does not necessarily improve performance  as the training can become very difficult due to the problem of vanishing gradients. In this work, we apply the the deep residual learning framework proposed by He et al. \cite{kmhe15}. It manages to learn the residual mapping of a few stacked layers to avoid the vanishing gradients problem.

\begin{figure*}
	\begin{center}
		\includegraphics[scale=.75]{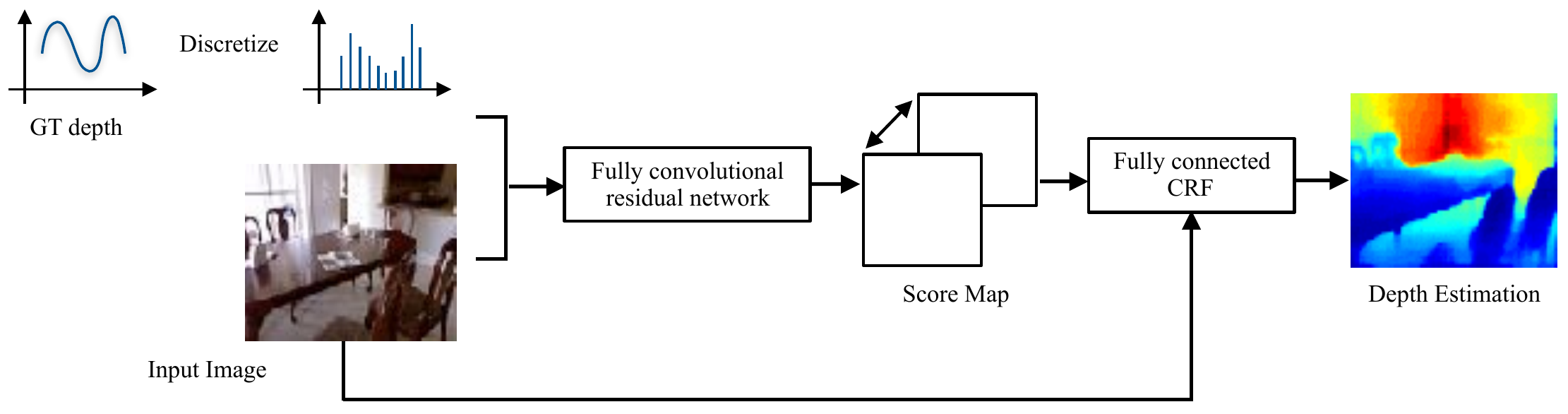}
	\end{center}
	\caption{An overview of our depth estimation model. It takes as input an image and output dense score maps. Fully-connected CRFs are then applied to obtain the final depth estimation.}
	\label{fig:overview}
\end{figure*}

An overview of our proposed depth estimation model is illustrated in Fig.~\ref{fig:overview}. It takes as input an arbitrarily sized image and outputs a dense score map. Fully connected CRFs are then applied to obtain the final depth estimation. The remaining content of the paper is organized as follows. Section \ref{sec:related} reviews some relevant work. Then we present the proposed method in Section \ref{sec:method}.  Experiment results are presented in Section \ref{sec:exp}. Finally,  Section \ref{sec:con} concludes the paper.

\section{Related Work}
\label{sec:related}

Previous  depth estimation methods are mainly based on geometric models. For example, the works of
\cite{Hedau,NIPS2010_4120,SchwingECCV2012} rely on box-shaped models and try to fit the box edges to those observed in the image.
These methods are limited to only model particular scene structures and therefore are not applicable for general-scene depth estimations.
More recently, non-parametric methods \cite{KKarsch} are explored. These methods consist of candidate images retrieval, scene alignment and then depth inference using optimizations with smoothness constraints.
These methods are based on the assumption that scenes with semantically similar appearances should have similar depth distributions when densely aligned.

Other methods attempt to exploit additional information. To name a few,
the authors of \cite{RussellT09} estimated depths through user annotations. The work of
\cite{LiuCVPR2010} performed semantic label prediction before depth estimation. The works of
\cite{Ladicky_2014_CVPR,Wang_2015_CVPR} have shown that jointly perform depth estimation and semantic labelling can help
each other. Given the fact that the extra source of information is not always available, most of recent works formulated depth estimation as a Markov Random Field (MRF) \cite{Saxena2005,Saxena2009,Saxena073-ddepth} or Conditional Random Field (CRF) \cite{Liu_2014_CVPR} learning problem.
These methods managed to  learn the parameters of MRF/CRF  in a supervised fashion
from a training set of monocular images and their corresponding ground-truth depth images.
The depth estimation problem then is formulated as a maximum a posteriori (MAP) inference problem on the CRF model.

With the popularity of deep convolutional neural networks (CNN) since the work of \cite{NIPS2012_4824}, some works attempted to solve the depth estimation problem using deep convolutional networks and achieved outstanding performance. Eigen et al. \cite{Eigen15} proposed a multi-scale architecture for predicting depths, surface normals and semantic labels. The multi-scale architecture is able to capture many image details without any superpixels or low-level segmentation. Liu et al. \cite{LiuSLR15} presented a deep convolutional neural field model for depth estimation. It learned the unary and pairwise potentials of continuous CRF in a unified deep network.
The model is based on fully convolutional networks (FCN) with a novel superpixel pooling method. Similarly, Li et al. \cite{LiB15} and Wang et al. \cite{Wang_2015_CVPR} also combined the CNNs with CRFs, they formulated depth estimation in a two-layer hierarchical CRF to enforce synergy between global and local predictions.

Anirban et al. \cite{RoyT16} proposed a neural regression forest (NRF) architecture which combines convolutional neural networks with random forests for predicting depths in the continuous domain via regression. The NRF processes a data sample with an ensemble of binary regression trees and the final depth estimation is made by fusing the individual regression results. It allows for parallelizable training of all shallow CNNs, and efficient enforcing of smoothness in depth estimation results. Laina et al. \cite{laina2016deeper} applied the deep residual networks for depth estimation. In order to improve the output resolution, they presented a novel way to efficiently learn feature map up-sampling within the network. They also presented a reverse Huber loss which is driven by the value distributions commonly present in depth maps for the network optimization.

Experiment results in the aforementioned works reveal that depth estimation benefits from: (a) an
increased number of layers in deep networks; (b) obtaining fine-level details. In this work,
we take advantage of the successful deep residual networks \cite{kmhe15} and formulate depth estimation as a dense prediction task.
We also apply fully connected CRFs \cite{phil2011} as post-processing. Although Laina et al. \cite{laina2016deeper}  also applied the deep residual network for depth estimation, our method is different from \cite{laina2016deeper} in 3 distinct ways: Firstly, we formulate depth estimation as a classification task, while \cite{laina2016deeper} formulated depth estimation as a regression task. Secondly, we can obtain the confidence of depth predictions which can be used during training and post-processing. Lastly, in order to obtain high resolution predictions, \cite{laina2016deeper} applied an up-sampling scheme while we simply use bilinear interpolation.

The aforementioned CNN based methods formulate depth estimation as a structured regression task due to the continuous property of depth values. However for different pixels in a single monocular image, the possible depth values have different distributions.
Depth values of some pixels are easy to predict while others are not.
The output of continuous regression lacks this confidence. In \cite{deepak16}, Pathak et al. presented a novel structured regression framework for image decomposition. It applied special constraints on the output space to capture the confidence of predictions. In \cite{bayesiansegnet}, Kendall et al. proposed a Bayesian neural network for semantic segmentation. It applied the Monte-Carlo dropout during training and obtained the confidence of predictions by multiple forward passes during evaluation. In this work, we obtain the confidence by simply formulating depth estimation as a classification task.

\section{Proposed Method}\label{sec:method}

In this section, we describe our depth estimation method in detail. We first introduce the network architecture, followed by the introduction of our loss function. Finally, we introduce the fully connected conditional random field (CRF) which is applied as post-processing.

\subsection{Network architecture}
We formulate our depth estimation as a spatially dense prediction task. When applying CNNs to this type of task, the input image is inevitably down-sampled due to the repeated combination of max-pooling and striding. In order to handle this, we follow the fully convolutional network (FCN) which has been proven to be successful in dense pixel labeling. It replaces the fully connected layers in conventional CNN architectures with convolutional layers.
By doing this, it makes the fully convolutional networks capable of taking input of arbitrarily sized images and output a down-sampled prediction map.
After applying  a simple upsample such as bilinear interpolation, the prediction map is of the same size of the input image.

The depth of CNN architectures is of great importance. Much recent works reveal that the VGG \cite{Simonyan14c} network outperforms the shallower AlexNet \cite{NIPS2012_4824}. However, simply stacking more layers to existing CNN architectures does not necessarily improve performance due to the notorious problem of vanishing gradients, which hampers convergence from the beginning during training. The recent
ResNet model solves this problem by adding  skip connections.
We follow the recent success of deep residual network with up to 152 layers \cite{kmhe15}, which is about
8$\times$ deeper than the VGG network but still having fewer parameters to optimize.

\begin{figure}
	\begin{center}
		\includegraphics[scale=.41]{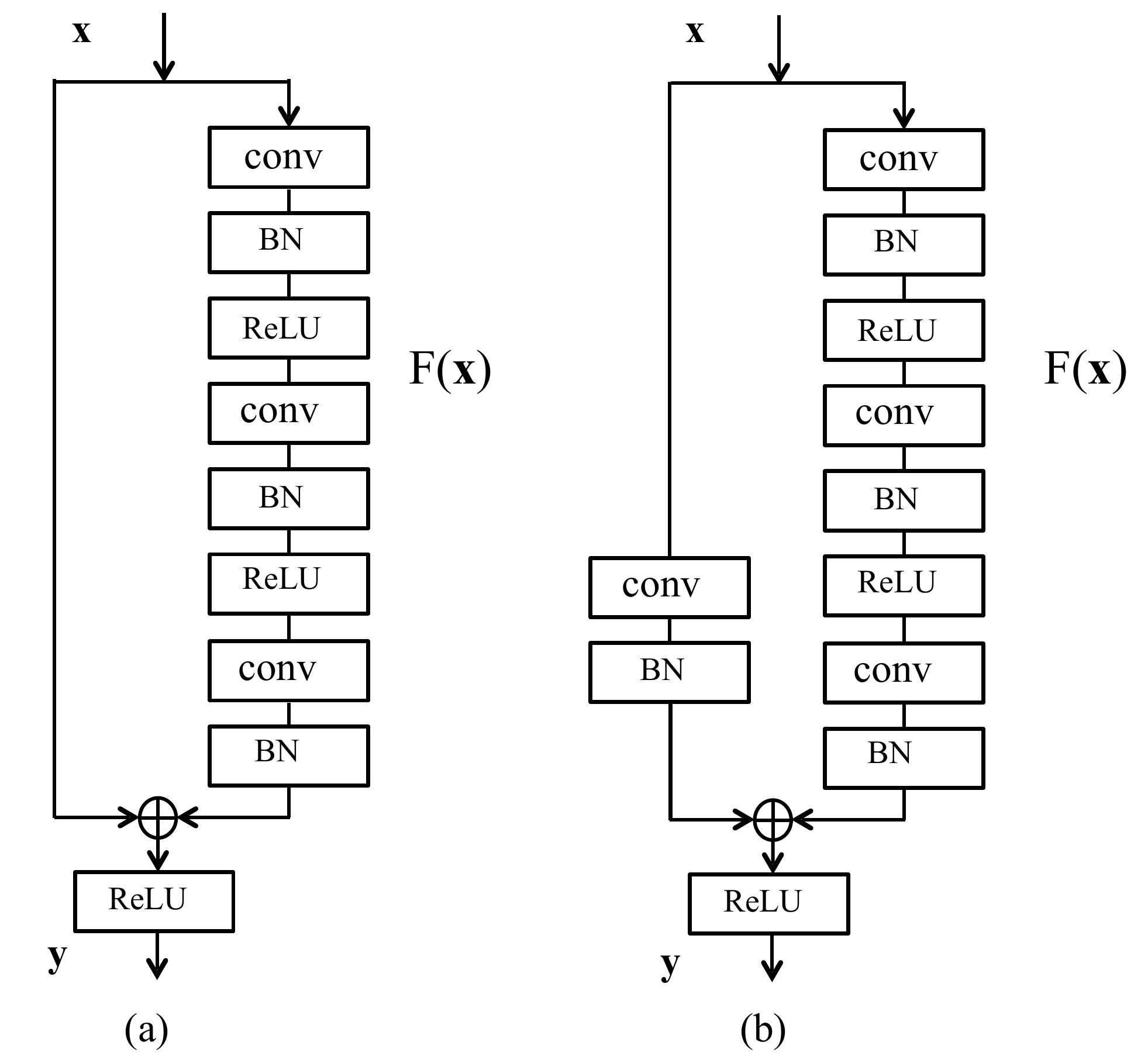}
	\end{center}
	\caption{Two types of building blocks that can be used in our depth estimation model. (a) building block with identity mapping. (b) building block with linear projection.}
	\label{fig:buildingblocks}
\end{figure}

Instead of directly learning the underlying mapping of a few stacked layers, the deep residual network learns the residual mapping. Then the original mapping can be realized by feedforward neural networks with ``shortcut connections''. Shortcut connections are those skipping one or more layers. In our model, we consider two shortcut connections and the building blocks are shown in Fig. \ref{fig:buildingblocks}. The building block illustrated in Fig. \ref{fig:buildingblocks}(a) is defined as:

\begin{equation} \label{block1}
\mathbf{y} = \mathnormal{F}(\mathbf{x},\{W_i\})+\mathbf{x},
\end{equation}
where $\mathbf{x}$ and $\mathbf{y}$ are the input and output matrices of stacked layers respectively. The function $\mathnormal{F}(\mathbf{x},\{W_i\})$ is the residual mapping that need to be learned. Since the shortcut connection is an element-wise addition, the dimensions of $\mathbf{x}$ and $\mathnormal{F}$ need to be same.

The building block illustrated in Fig. \ref{fig:buildingblocks}(b) is defined as:

\begin{equation}
\mathbf{y} = \mathnormal{F}(\mathbf{x},\{W_i\})+W_s\mathbf{x}.
\end{equation}
Compared to the shortcut connection in Eq.~(\ref{block1}), a linear projection $\mathnormal{W_s}$ is applied to match the dimensions of $\mathbf{x}$ and $\mathnormal{F}$.

\begin{figure*}
	\begin{center}
		\includegraphics[scale=.685]{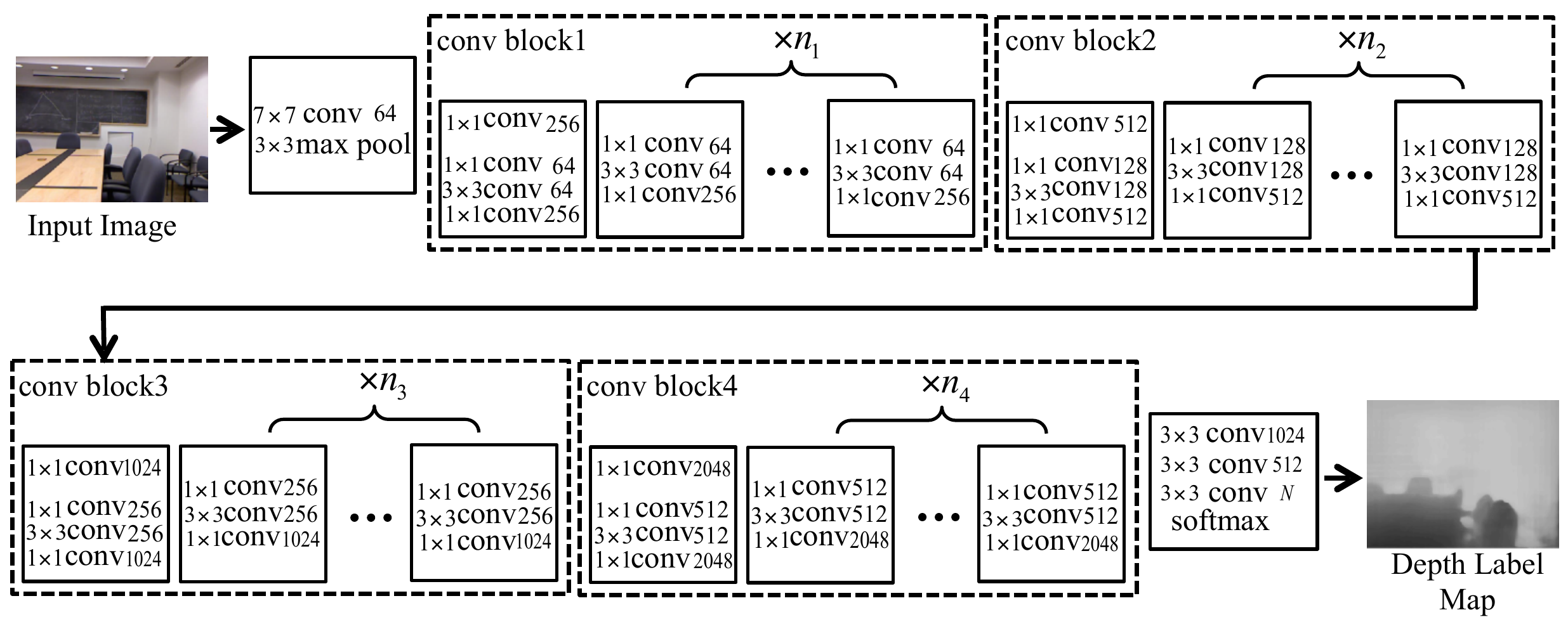}
	\end{center}
	\caption{Network architecture of our depth estimation model. The input image is fed into a convolutional layer, a max pooling layer and 4 convolution blocks. We consider network architectures   with 101 and 152 layers. The value of $[n_1,n_2,n_3,n_4]$ is $[2,3,22,2]$ for the
    101-layer network architecture and $[2,7,35,2]$ for the 152-layer network architecture. The last 4 layers are 3 convolutional layers and a softmax layer. The output map is downsampled by a factor of 8 and we preform bilinear interpolation during prediction.}
	\label{fig:network}
\end{figure*}

The overall network architecture of our depth estimation model is illustrated in Fig. \ref{fig:network}. The input image is fed into a convolutional layer, a max pooling layer followed by 4 convolution blocks. Each convolution block starts with a building block with linear projection followed by different numbers
of building blocks with identity mapping.
In this article, we consider two deep residual network architectures with 101 and 152 layers respectively. For the network architecture with 101 layers, the number of building blocks with identity mapping in the four convolution blocks (i.e., $n_1,n_2,n_3,n_4$ in Fig. \ref{fig:network}) are 2, 3, 22 and 2 respectively. As for the network architecture with 152 layers, the numbers are 2, 7, 35 and 2. The last four layers are three convolutional layers with channels 1024,512 and $N$, and a softmax layer, where $N$ is the number of ground-truth labels. Batch normalization and ReLU layers are performed between these convolutional layers. Downsampling is performed by pooling or convolutional layers that have a stride of 2. These include the first $7\times7$ convolutional layer, the first $3\times3$ max pooling layer, and the first building block of convolution block 2 in Fig.~\ref{fig:network}. As a result, the output prediction map is downsampled by a factor of 8. During prediction, we perform a bilinear interpolation on this map to make it the same size with the input image.

\subsection{Loss function}
In this work,  we use the pixel-wise multinomial logistic loss function as we formulate depth estimation as a classification task. We uniformly discretize the continuous depth values into multiple bins in the log space. Each bin covers a range of depth values and we label the bins according to the range (i.e., the label index of a pixel indicates its distance). The depth labels however are different from the labels of typical classification tasks. For typical classification tasks such as semantic segmentation and object detection, the predictions that are different from ground-truth labels are considered wrong and contribute nothing in updating network parameters. As for depth estimation, the predictions that are close to ground-truth depth labels can also help in updating network parameters. This is achieved by an ``information gain" matrix in our loss function.

Specifically, our loss function is defined as:

\begin{equation}\label{lossfunc}
\mathnormal{L} = -\frac{1}{N}
\sum_{i=1}^{N}\sum_{D=1}^{B}
H(D_{i}^{*},D)\log(\mathnormal{P(D|z_i)}),
\end{equation}
where $\mathnormal{D_{i}^{*}}\in[1,\dots,B]$ is the ground-truth depth label of pixel $\mathnormal{i}$ and $B$ is the total number of discretization bins. $\mathnormal{P(D|z_i)} = {e^{z_i,D}}/{\sum_{d=1}^{B}{e^{z_{i,d}}}}$ is the probability of pixel $\mathnormal{i}$ labelled with $\mathnormal{D}$. $\mathnormal{z_{i,d}}$ is the output of the last convolutional layer in the network. The ``information gain" matrix $H$ is a $B \times B$ symmetric matrix with elements $H(p,q) = \exp[-\alpha(p-q)^2]$ and $\alpha$ is a constant. It encourages the predicted depth labels that are closer to ground-truths have higher contributions in updating network parameters.

During prediction, we set the depth value of each pixel to be the center of its corresponding bin. By formulating depth estimation as classification, we can get the confidence of each prediction in the form of probability distribution. This confidence can also be applied during post-processing via fully connected CRFs.

\subsection{Fully connected conditional random fields}

A deep convolutional network typically does  not explicitly take the dependency among local variables into consideration.
It does so only implicitly through the field of view. That is why the size of field of view is important in terms of the performance of a CNN.
In order to greatly refine the network output, we apply the fully connected CRF proposed in \cite{phil2011} as post-processing. It connects all pairs of individual pixels in the image. Specifically, the energy function of a fully connected CRF is the sum of unary potential $\mathnormal{U}$ and pairwise potential $\mathnormal{V}$:

\begin{equation}
\mathnormal{E(\mathbf{D})} = \sum_{i}\mathnormal{U}(\mathnormal{D_i}) + \sum_{i,j}\mathnormal{V}(\mathnormal{D_i,D_j}),
\end{equation}
where $\mathbf{D}$ is the predicted depth labels of pixels and $i$, $j$ are pixel indices. We use the logistic loss of pixel defined in Eq. (\ref{lossfunc}) as the unary potential, which is
\[
 \mathnormal{U}(\mathnormal{D_i}) = \mathnormal{L(D_i)} = -\log(\mathnormal{P(D_{i}|z_i)}).
 \]
 The pairwise potential is defined as
 \[
     \sum_{i,j}\mathnormal{V}(\mathnormal{D_i,D_j}) = \Delta(D_i,D_j)\sum_{s=1}^{M}w_s \cdot k^{s}(\mathbf{f}_i,\mathbf{f}_j),
     \]
 where $\Delta(D_i,D_j)$ is a penalty term on the labelling. Since the label here indicates depth, we enforce a relatively larger penalty for labellings that are far away from ground-truth. For simplicity, we use the absolute difference between two label values to be the penalty: $\Delta(D_i,D_j) = |D_i-D_j|$. There is one pairwise term for each pair of pixels in the image no matter how far they are from each other (i.e., the model's factor graph is fully connected).

Each $k^s$ is the Gaussian kernel depends on features (denoted as $\mathbf{f}$) extracted for pixel $i$ and $j$ and is weighted by parameter $w_s$. Following \cite{phil2011}, we adopt bilateral positions and color terms, specifically, the kernels are:

\begin{equation}
\mathnormal{w_1}\exp(-\frac{\|p_i-p_j\|^2}{2\sigma_{\alpha}^{2}}-\frac{\|I_i-I_j\|^2}{2\sigma_{\beta}^{2}})
    + \mathnormal{w_2}\exp(-\frac{\|p_i-p_j\|^2}{2\sigma_{\gamma}^{2}}).
\end{equation}
The first kernel is appearance kernel, which
depends on both pixel positions (denoted as $p$) and pixel color intensities (denoted as $I$). It is inspired by the observation that nearby pixels with similar color are likely to be in the same depth range. The degrees of nearness and similarity are controlled by hyper parameters $\sigma_{\alpha}$ and $\sigma_{\beta}$. The second kernel is smoothness kernel which removes small isolated regions, the scale of smoothness is controlled by $\sigma_{\gamma}$.

\section{Experiments}
\label{sec:exp}
We evaluate our proposed depth estimation approach on 2 benchmark RGB-D datasets: the indoor NYUD2 \cite{Silberman:ECCV12} dataset and the outdoor KITTI \cite{Geiger2013IJRR} dataset. We organize our experiments into the following three parts:

(1) We show the effectiveness of our depth discretization scheme and compare our discrete depth label classification with continuous depth value regression.

(2) We evaluate the contribution of different components in our proposed approach.

(3) We compare our proposed approach with state-of-the-art methods to show that our approach performs better in both indoor and outdoor scenes. Several measures commonly used in prior works are applied for quantitative evaluations:

$\bullet$ root mean squared error (rms): $\sqrt{\frac{1}{T}\sum_{p}(d_{gt}-d_p)^2}$

$\bullet$ average relative error (rel): $\frac{1}{T}\sum_{p}\frac{|d_{gt}-d_{p}|}{d_{gt}}$

$\bullet$ average $\log_{10}$ error (log10): $\frac{1}{T}\sum_{p}|\log_{10}d_{gt} - \log_{10}d_p|$

$\bullet$ root mean squared log error (rmslog) $\sqrt{\frac{1}{T}\sum_{p}(\log d_{gt} - \log d_p)^2}$

$\bullet$ accuracy with threshold $thr$:

percentage ($\%$) of $d_p$ s.t. $\max(\frac{d_{gt}}{d_p},\frac{d_p}{d_{gt}}) = \delta < thr$
where $d_{gt}$ and $d_p$ are the ground-truth and predicted depths respectively of pixels, and $T$ is the total number of pixels in all the evaluated images.

\subsection{Depth label classification vs. depth value regression}
\begin{figure}
	\begin{center}
		\includegraphics[scale=.62]{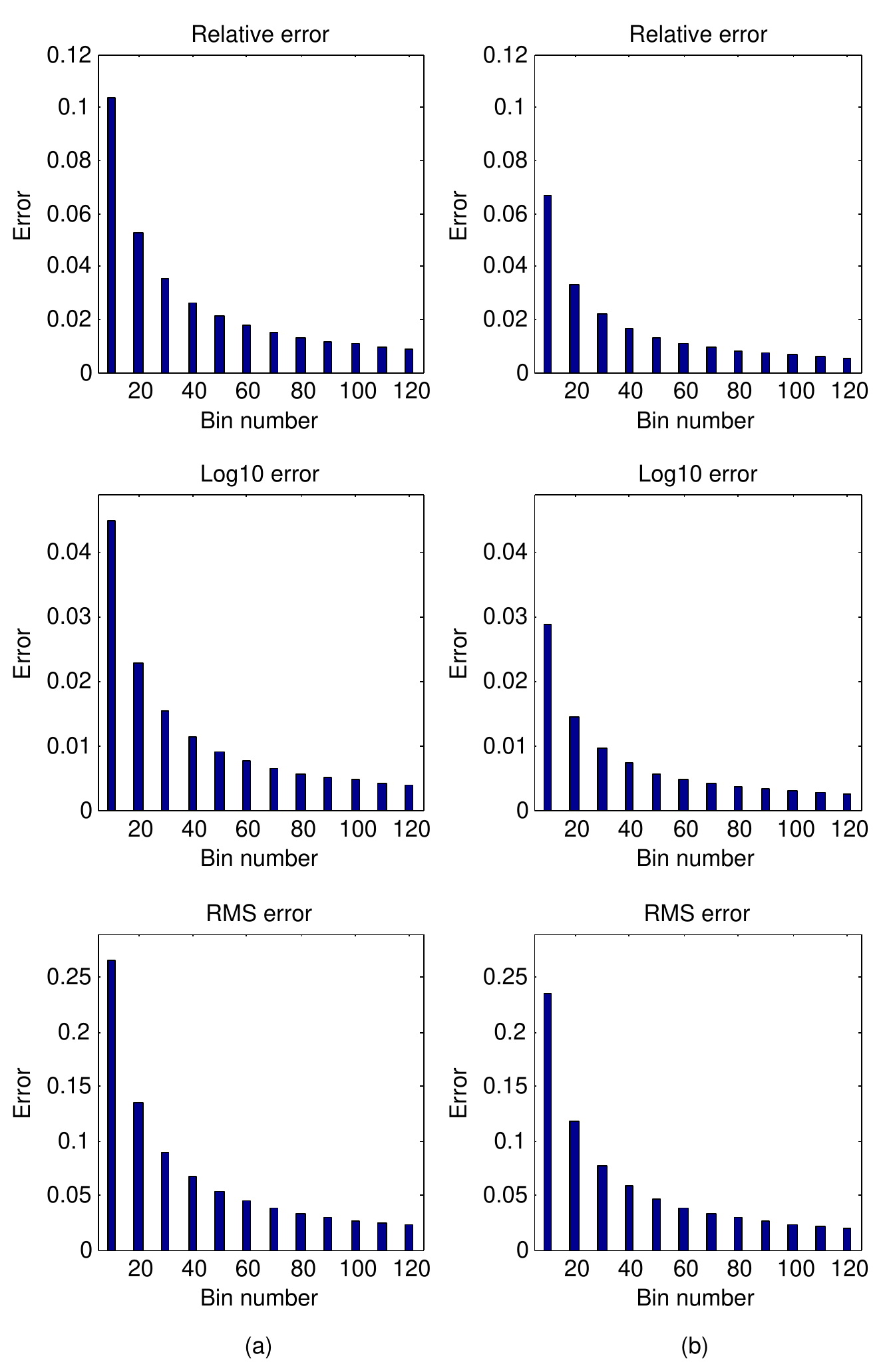}
	\end{center}
	\caption{Quantitative evaluations of discretized ground-truth depth values of the NYUD2 dataset. (a): errors of ground-truth depth values discretized in linear space. (b): errors of ground-truth depth values discretized in the log space.}
	\label{fig:gt_validate}
\end{figure}

Discretizing continuous data would inevitably discard some information. In this part, we first show that the discretization of continuous depth values degrades the depth estimation model by negligible amount. Specifically, we equally discretize the ground-truth depth values of test images in the NYUD2 dataset into different numbers of bins in the linear and log space respectively and calculate three errors as is mentioned above. The results are illustrated in Fig. \ref{fig:gt_validate}.

We can see from Fig. \ref{fig:gt_validate} that with the increment of discretization bins, the errors of discretized ground-truth depths decrease and stop at a negligible amount. And the discretization in the log space leads to lower error than the discretization in the linear space. 

As for the accuracies, all the discretized ground-truth depths can reach 100\% except for the accuracy with threshold 1.25 when linearly discretizing the ground-truth depths into 10 bins. From this experiment we can see that converting the ground-truth depths from continuous values to discrete labels has negligible effect on the performance. We can reformulate depth estimation from a conventional regression task to a classification task.

\begin{table}
	\caption{Depth estimation results by continuous depth value regression and discrete depth label classification for the NYUD2 and KITTI datasets. The first row is the result by regression. The following rows are results of depth label classification with different number of discretization bins.}
	\centering
	\label{table:regress_n_classify}
	\begin{tabular}{@{\hskip 0.05cm}c@{\hskip 0.07cm}c@{\hskip 0.07cm}c@{\hskip 0.07cm}c @{\hskip 0.39cm}c@{\hskip 0.15cm}c@{\hskip 0.15cm}c}
        \noalign{\smallskip}
		\hline
		\noalign{\smallskip}
						& \multicolumn{3}{c}{\small Accuracy}    & \multicolumn{3}{c}{\small Error}  \\
		                 & \small $\delta<1.25$   &\small $\delta<1.25^2$    &\small $\delta<1.25^3$    &\small rel    &\small log10  &\small rms  \\
		\noalign{\smallskip}
		\hline
		\noalign{\smallskip}
		& \multicolumn{5}{c}{\small NYUD2}     \\
		\hline
		\noalign{\smallskip}
		\small Regression           &\small 65.3\%   & \small 91.5\%   & \small 97.4\%   & \small 0.231  &\small 0.095  &\small 0.778  \\
		\noalign{\smallskip}
		\small 10 bins              &\small 69.4\%   & \small \bf 92.4\%   & \small 97.5\%   & \small 0.213    &\small 0.091    &\small 0.754  \\
		\noalign{\smallskip}
		\small 30 bins              &\small 70.5\%   &\small 92.1\%   &\small \bf 97.8\%   &\small 0.210    &\small \bf 0.090    &\small 0.751  \\
		\noalign{\smallskip}
		\small 50 bins              &\small 68.9\%   &\small 91.9\%   &\small 97.0\%   &\small 0.209    &\small 0.092    &\small 0.750  \\
		\noalign{\smallskip}
		\small 80 bins              &\small \bf 70.6\%   &\small 92.0\%   &\small 97.6\%   &\small 0.211    &\small 0.091    &\small \bf 0.747   \\
		\noalign{\smallskip}
        \small 100 bins             &\small 70.1\%   &\small 92.1\%   &\small 97.3\%   &\small \bf 0.209    &\small 0.091    &\small 0.749  \\
		\hline
		\noalign{\smallskip}
		& \multicolumn{5}{c}{\small KITTI}     \\
		\hline
		\noalign{\smallskip}
		\small Regression           &\small 67.5\%   &\small 88.6\%   &\small 90.4\%   &\small  0.279    &\small 0.104    &\small 7.916  \\
		\noalign{\smallskip}
		\small 50 bins            &\small 76.3\%   &\small \bf 92.1\%   &\small  96.3\%   &\small  0.183    &\small  0.077  &\small \bf 6.209  \\
		\noalign{\smallskip}
		\small 80 bins              &\small \bf 77.1\%  &\small 91.7\%   &\small 96.6\%   &\small \bf 0.180    &\small \bf 0.072  &\small 6.311  \\
		\noalign{\smallskip}
		\small 120 bins             &\small 76.8\%   &\small 91.9\%   &\small \bf 96.7\%   &\small 0.187    &\small 0.076   &\small 6.263  \\
		\hline
		\noalign{\smallskip}
		
	\end{tabular}
\end{table}

We next compare our proposed depth estimation by classification with the conventional depth regression and show the results in Table \ref{table:regress_n_classify}. In this experiment, we apply the deep residual network with 101 layers and the parameters are initialized with the ResNet101 model in \cite{kmhe15} which is trained on the ImageNet classification dataset. We train our models on standard NYUD2 training set with 795 images and standard KITTI training set with 700 images \cite{EigenPF14} for fast comparison. As for the test sets, we select 650 and 700 images from the raw NYUD2 and KITTI test sets respectively as validation sets. For depth regression, the loss function is standard $L2$ norm which minimizes the squared euclidean norm between predicted and ground-truth depths. The output depth map is upsampled to the same size of the input image through bilinear interpolation. As for our depth estimation by classification, we discretize the continuous depth values into different numbers of bins in the log space. We do not apply CRF post-precessing for both regression and classification. As we can see from Table \ref{table:regress_n_classify} that depth estimation by classification outperforms the conventional depth regression, and the performance of depth classification is not very sensitive to the number of discretization bins.

One important reason for depth estimation by classification outperforms the depth regression is that the regression tends to converge to the mean depth values. This may cause larger errors in areas that are either too far from or too close to the camera. The classification with the information gain may alleviate this problem. In order to testify this, we break down the NYUD2 ground-truth depths into 3 ranges and report the results in Table \ref{table:regionGT}. The general setting is the same with the aforementioned experiment. The ground-truth depths are discretized into 100 bins in the log space and the $\alpha$ defined in Eq.~(\ref{lossfunc}) is set to 0.2.

\begin{table}
	\caption{Test results on the NYUD2 dataset with different ground-truth ranges. We break down the ground-truth depths into 0m-3m, 3m-7m and 7m-10m.}
	\centering
	\label{table:regionGT}
	\begin{tabular}{@{\hskip 0.05cm}c@{\hskip 0.07cm}c@{\hskip 0.07cm}c@{\hskip 0.07cm}c @{\hskip 0.39cm}c@{\hskip 0.15cm}c@{\hskip 0.15cm}c}
        \noalign{\smallskip}
		\hline
		\noalign{\smallskip}
						& \multicolumn{3}{c}{\small Accuracy}    & \multicolumn{3}{c}{\small Error}  \\
		                 & \small $\delta<1.25$   &\small $\delta<1.25^2$    &\small $\delta<1.25^3$    &\small rel    &\small log10  &\small rms  \\
		\noalign{\smallskip}
		\hline
		\noalign{\smallskip}
		& \multicolumn{5}{c}{\small Regression}     \\
		\hline
		\noalign{\smallskip}
		\small 0m-3m           & \small 65.7\%   & \small 90.9\%   & \small 97.4\%   & \small 0.233  &\small 0.087  &\small 0.561  \\
		\noalign{\smallskip}
		\small 3m-7m           & \small 70.3\%   & \small 95.5\%   & \small 99.5\%   & \small 0.175  &\small 0.075  &\small 0.936  \\
		\noalign{\smallskip}
		\small 7m-10m          & \small 45.0\%   & \small 75.4\%   & \small 93.5\%   & \small 0.242  &\small 0.129  &\small 2.346  \\
		\hline
		\noalign{\smallskip}
		& \multicolumn{5}{c}{\small Classification}     \\
		\hline
		\noalign{\smallskip}
		\small 0m-3m           &\small 69.6\%   &\small 91.2\%   &\small 97.2\%   &\small  0.216    &\small 0.083    &\small 0.561  \\
		\noalign{\smallskip}
		\small 3m-7m           &\small 76.0\%   &\small 94.9\%   &\small 98.6\%   &\small  0.151    &\small 0.070    &\small 0.857  \\
		\noalign{\smallskip}
		\small 7m-10m          &\small 49.7\%   &\small 74.9\%   &\small 93.1\%   &\small  0.238    &\small 0.126    &\small 2.199  \\
		\hline
		\noalign{\smallskip}
		
	\end{tabular}
\end{table}

\subsection{Component evaluation}
In this section, we analyze the contribution of key components including the information gain matrix, fully connected CRFs and network architectures in our proposed approach. We evaluate depth estimation on both the NYUD2 and KITTI datasets. We use the standard training set containing 795 images of the NYUD2 dataset and evaluate on the standard 654 test images. The continuous depth values are discretized into 100 bins in the log space. As for the KITTI dataset, we apply the same split in \cite{EigenPF14} which contains 700 training images and 697 test images. We only use left images and discretize the continuous depth values into 50 bins in the log space. We cap the maximum depth to be 80 meters. During training, we ignore the missing values in ground-truth depths and only evaluate on valid points.

\subsubsection{Benefit of information gain matrix}
In this part, we evaluate the contribution of the information gain matrix in our loss function. We train the ResNet101 model on both the NYUD2 and KITTI datasets with and without information gain matrices. The $\alpha$ defined in Eq.~(\ref{lossfunc}) is set to 0.2 and 0.5 for NYUD2 and KITTI respectively. In our experiments, we find that the performance is not sensitive to $\alpha$. The results are illustrated in Table \ref{table:info_gain}. As we can see from this table that the information gain matrix improves the performance of both indoor and outdoor depth estimation. 

\begin{table}
	\renewcommand\arraystretch{0.65}
	\caption{Test results on the NYUD2 and KITTI datasets with and without information gain matrices. For each dataset, the first row is the result without information gain matrix, the second row is the result with information gain matrix.}
	\centering
	\label{table:info_gain}
	\begin{tabular}{@{\hskip 0.05cm}c@{\hskip 0.07cm}c@{\hskip 0.07cm}c@{\hskip 0.07cm}c @{\hskip 0.39cm}c@{\hskip 0.15cm}c@{\hskip 0.15cm}c}
        \noalign{\smallskip}
		\hline
		\noalign{\smallskip}
						& \multicolumn{3}{c}{\small Accuracy}    & \multicolumn{3}{c}{\small Error}  \\
		                 & \small $\delta<1.25$   &\small $\delta<1.25^2$    &\small $\delta<1.25^3$    &\small rel    &\small log10  &\small rms  \\
		\noalign{\smallskip}
		\hline
		\noalign{\smallskip}
		& \multicolumn{5}{c}{\small NYUD2}     \\
		\hline
		\noalign{\smallskip}
   \small Plain    &\small 70.9\%   &\small 92.1\%   &\small 98.0\%   &\small 0.193    &\small 0.079    &\small 0.716  \\
		\noalign{\smallskip}
   \small Infogain   &\small \bf 72.2\%   &\small \bf 92.6\%   &\small \bf 98.0\%   &\small \bf 0.192    &\small \bf 0.077    &\small \bf 0.688  \\	
		\hline
		\noalign{\smallskip}
		& \multicolumn{5}{c}{\small KITTI}     \\
		\hline
		\noalign{\smallskip}
   \small Plain    &\small 79.9\%   &\small 93.7\%   &\small 97.6\%   &\small 0.166    &\small 0.067    &\small 5.443  \\
		\noalign{\smallskip}
   \small Infogain   &\small \bf 81.4\%   &\small \bf 93.9\%   &\small \bf 97.6\%   &\small \bf 0.153    &\small \bf 0.062    &\small \bf 5.290  \\	
		\hline
		\noalign{\smallskip}
		
	\end{tabular}
\end{table}

\subsubsection{Benefit of fully connected CRFs}
In order to evaluate the effect of the fully connected CRFs, we first train the ResNet101 model on both the NYUD2 and KITTI datasets, and then apply the fully connected CRFs as post-processing.  We illustrate the results in Table \ref{table:fc_CRF}. As we can see from the table, the fully-connected CRF can improve the depth estimation of both indoor and outdoor scenes.

\begin{table}
	\renewcommand\arraystretch{0.65}
	\caption{Test results on the NYUD2 and KITTI datasets with and without the fully connected CRFs as post-processing. For each dataset, the first row is the result without CRFs, the following row is the result with CRFs.}
	\centering
	\label{table:fc_CRF}
	\begin{tabular}{@{\hskip 0.05cm}c@{\hskip 0.07cm}c@{\hskip 0.07cm}c@{\hskip 0.07cm}c @{\hskip 0.39cm}c@{\hskip 0.15cm}c@{\hskip 0.15cm}c}
        \noalign{\smallskip}
		\hline
		\noalign{\smallskip}
						& \multicolumn{3}{c}{\small Accuracy}    & \multicolumn{3}{c}{\small Error}  \\
		                 & \small $\delta<1.25$   &\small $\delta<1.25^2$    &\small $\delta<1.25^3$    &\small rel    &\small log10  &\small rms  \\
		\noalign{\smallskip}
		\hline
		\noalign{\smallskip}
		& \multicolumn{5}{c}{\small NYUD2}     \\
		\hline
		\noalign{\smallskip}
   \small Plain    &\small 70.9\%   &\small \bf 92.1\%   &\small 98.0\%   &\small 0.193    &\small 0.079    &\small 0.716  \\
		\noalign{\smallskip}
   \small CRF     &\small \bf 71.3\%   &\small 92.0\%   &\small \bf 98.0\%   &\small \bf 0.190    &\small \bf 0.079    &\small \bf 0.696 \\
		\noalign{\smallskip}
		\hline
		\noalign{\smallskip}
		& \multicolumn{5}{c}{\small KITTI}     \\
		\hline
		\noalign{\smallskip}
   \small Plain    &\small 79.9\%   &\small 93.7\%   &\small 97.6\%   &\small \bf 0.166    &\small 0.067    &\small 5.443  \\
		\noalign{\smallskip}
   \small CRF   &\small \bf 81.0\%   &\small \bf 94.1\%   &\small \bf 97.9\%   &\small 0.167    &\small \bf 0.066    &\small \bf 5.349 \\
		\hline
		\noalign{\smallskip}
		
	\end{tabular}
\end{table}

\begin{table}
	\renewcommand\arraystretch{0.65}
	\caption{Test results on the NYUD2 dataset with different network structures. The first row is the result of the VGG16 net, the following two rows are the results of deep residual networks. We also show the total number of parameters of the three networks in the last row.}
	\centering
	\label{table:networks}
	\begin{tabular}{@{\hskip 0.05cm}c@{\hskip 0.07cm}c@{\hskip 0.07cm}c@{\hskip 0.07cm}c @{\hskip 0.39cm}c@{\hskip 0.15cm}c@{\hskip 0.15cm}c}
        \noalign{\smallskip}
		\hline
		\noalign{\smallskip}
						& \multicolumn{3}{c}{\small Accuracy}    & \multicolumn{3}{c}{\small Error}  \\
		                 & \small $\delta<1.25$   &\small $\delta<1.25^2$    &\small $\delta<1.25^3$    &\small rel    &\small log10  &\small rms  \\
		\noalign{\smallskip}
		\hline
		\noalign{\smallskip}
   \small VGG16     &\small  62.1\%   &\small 87.2\%   &\small 96.0\%   &\small 0.236    &\small  0.097  &\small  0.857 \\
   		\noalign{\smallskip}
   \small ResNet101    &\small  70.9\%   &\small  92.1\%   &\small  98.0\%   &\small  0.193    &\small  0.079  &\small  0.716  \\
		\noalign{\smallskip}
   \small ResNet152     &\small \bf 71.2\%   &\small \bf 92.3\%   &\small \bf 98.0\%   &\small \bf 0.187    &\small \bf 0.071    &\small \bf 0.681 \\
		\noalign{\smallskip}
		\hline
		
		\noalign{\smallskip}
		& \multicolumn{2}{c}{\small VGG16}  & \multicolumn{2}{c}{\small ResNet101}  & \multicolumn{2}{c}{\small ResNet152}\\
		\noalign{\smallskip}

		 \small Parameters   & \multicolumn{2}{c}{\small $13.9\times10^{7}$}  & \multicolumn{2}{c}{\small $6.7\times10^{7}$}  & \multicolumn{2}{c}{\small $8.2\times10^{7}$}\\
										
		\hline
		
	\end{tabular}
\end{table}

\begin{table*}
	\renewcommand\arraystretch{0.65}
	\caption{Comparison with state-of-the-art on the NYUD2 dataset. The first 4 rows are results by recent depth estimation models. The last row is the result of our approach.}
	\centering
	\label{table:state-of-art}
	\begin{tabular}{@{\hskip 0.1cm}c@{\hskip 0.2cm}c@{\hskip 0.2cm}c@{\hskip 0.2cm}c @{\hskip 0.6cm}c@{\hskip 0.25cm}c@{\hskip 0.25cm}c}
        \noalign{\smallskip}
		\hline
		\noalign{\smallskip}
						& \multicolumn{3}{c}{\small Accuracy}    & \multicolumn{3}{c}{\small Error}  \\
		                 &\small $\delta<1.25$   &\small $\delta<1.25^2$    &\small $\delta<1.25^3$    &\small rel    &\small log10     &\small rms  \\
		\noalign{\smallskip}
		\hline
		\noalign{\smallskip}
  \small  Wang et al. \cite{Wang_2015_CVPR}  & \small 60.5\%   &\small 89.0\%   &\small 97.0\%   &\small 0.210    &\small 0.094    &\small 0.745  \\
		\noalign{\smallskip}
  \small  Liu et al. \cite{LiuSLR15}         &\small 65.0\%   &\small 90.6\%   &\small 97.6\%   &\small 0.213    &\small 0.087    &\small 0.759  \\
    		\noalign{\smallskip}
  \small  Anirban et al. \cite{RoyT16}       & -   & -   & -   &\small 0.187    &\small 0.078    &\small 0.744  \\
		\noalign{\smallskip}
   \small Eigen et al. \cite{Eigen15}        &\small 76.9\%   &\small 95.0\%   &\small  98.8\%   &\small 0.158    &   -      &\small  0.641  \\
    		\noalign{\smallskip}
    \small	Laina et al. \cite{laina2016deeper}			&\small 81.1\%   &\small 95.3\%   &\small  98.8\%   &\small \bf 0.127    &\small \bf  0.055      &\small  0.573  \\
		\noalign{\smallskip}
   \small Ours    	&\small  \bf 81.9\%   &\small  \bf 96.5\%   &\small \bf 99.2\%   &\small  0.141    &\small  0.060    &\small \bf 0.540  \\
		\hline
	\end{tabular}
\end{table*}

\begin{table*}
	\renewcommand\arraystretch{0.65}
	\caption{Comparison with state-of-the-art results on the KITTI dataset. We cap the maximum depth to 50 and 80 meters to compare with recent works. For the work in~\cite{godard2016unsupervised}, we also report their results with additional training images in the CityScapes dataset~\cite{Cordts2016Cityscapes} and denote as Godard et al. CS.}
	\centering
	\label{table:stat-of-art_kitti}
	\begin{tabular}{@{\hskip 0.32cm}c@{\hskip 0.25cm}c@{\hskip 0.25cm}c@{\hskip 0.25cm}c @{\hskip 0.6cm}c@{\hskip 0.30cm}c@{\hskip 0.30cm}c}
        \noalign{\smallskip}
		\hline
		\noalign{\smallskip}
						& \multicolumn{3}{c}{\small Accuracy}    & \multicolumn{3}{c}{\small Error}  \\
		                &\small $\delta<1.25$   &\small $\delta<1.25^2$   &\small $\delta<1.25^3$  &\small rel  &\small rmslog  &\small rms  \\
		\noalign{\smallskip}
		\hline
		\noalign{\smallskip}
		& \multicolumn{5}{c}{\small Cap 80 meters}     \\
		\hline
		\noalign{\smallskip}
    \small Liu et al.~\cite{LiuSLR15}       &\small 65.6\%    &\small 88.1\%   &\small 95.8\%   &\small 0.217   &\small -   &\small 7.046  \\
		\noalign{\smallskip}
    \small Eigen et al.~\cite{EigenPF14}     &\small 69.2\%    &\small 89.9\%   &\small 96.7\%   &\small 0.190   &\small 0.270   &\small 7.156  \\
		\noalign{\smallskip}
    \small Godard et al.~\cite{godard2016unsupervised}    &\small 81.8\%    &\small 92.9\%   &\small 96.6\%   &\small 0.141   &\small 0.242   &\small 5.849  \\
		\noalign{\smallskip}
    \small Godard et al. CS~\cite{godard2016unsupervised}   &\small 83.6\%    &\small 93.5\%   &\small 96.8\%   &\small 0.136   &\small  0.236  &\small 5.763  \\
    		\noalign{\smallskip}
    \small  Ours     &\small \bf 88.7\%    &\small \bf 96.3\%   &\small \bf 98.2\%   &\small \bf 0.115   &\small \bf 0.198  &\small \bf 4.712  \\
		\hline
		\noalign{\smallskip}
		& \multicolumn{5}{c}{\small Cap 50 meters}     \\
		\hline
		\noalign{\smallskip}
    \small  Garg  et al.~\cite{garg2016unsupervised}   &\small 74.0\%    &\small 90.4\%   &\small 96.2\%   &\small 0.169   &\small  0.273  &\small 5.104  \\
		\noalign{\smallskip}
    \small  Godard et al.~\cite{godard2016unsupervised}   &\small 84.3\%    &\small 94.2\%   &\small 97.2\%   &\small 0.123   &\small  0.221  &\small 5.061  \\
    		\noalign{\smallskip}
    \small  Godard et al. CS~\cite{godard2016unsupervised}  &\small 85.8\%    &\small 94.7\%   &\small 97.4\%   &\small 0.118   &\small  0.215  &\small 4.941  \\
    		\noalign{\smallskip}
    \small  Ours  &\small \bf 89.8\%   &\small \bf 96.6\%   &\small \bf 98.4\%   &\small \bf 0.107   &\small \bf 0.187  &\small \bf 3.605  \\
		\hline
	\end{tabular}
\end{table*}

\subsubsection{Network Comparisons}
In this part, we compare the performance of deep residual networks with the baseline VGG16 net \cite{Simonyan14c} on the NYUD2 dataset. Since we formulate depth estimation as a classification task, we can apply network structures that perform well on semantic segmentation task. Specifically, for the VGG16 net, we apply the structure in \cite{LinSRH15}. We keep the layers up to ``fc6" in VGG16 net and add 2 convolutional layers with 512 channels, and 2 fully-connected layers with 512 and 100 channels respectively. The results are illustrated in Table~\ref{table:networks}. The performance of residual networks unsurprisingly outperform the VGG16 net, reinforcing the importance of network depth. Note that the performance by the ResNet152 improves little to the ResNet101, this is caused by the overfitting as the training set contains only 795 images. We also compare the number of parameters in the Table~\ref{table:networks}.

\subsection{State-of-the-art comparisons}
In this section, we evaluate our approach on the NYUD2 and KITTI datasets and compare with recent depth estimation methods. We apply the deep residual network with 152 layers and the parameters are initialized with the ResNet152 model in \cite{kmhe15}.

\subsubsection{NYUD2}
\begin{figure*}
	\begin{center}
		\includegraphics[scale=.765]{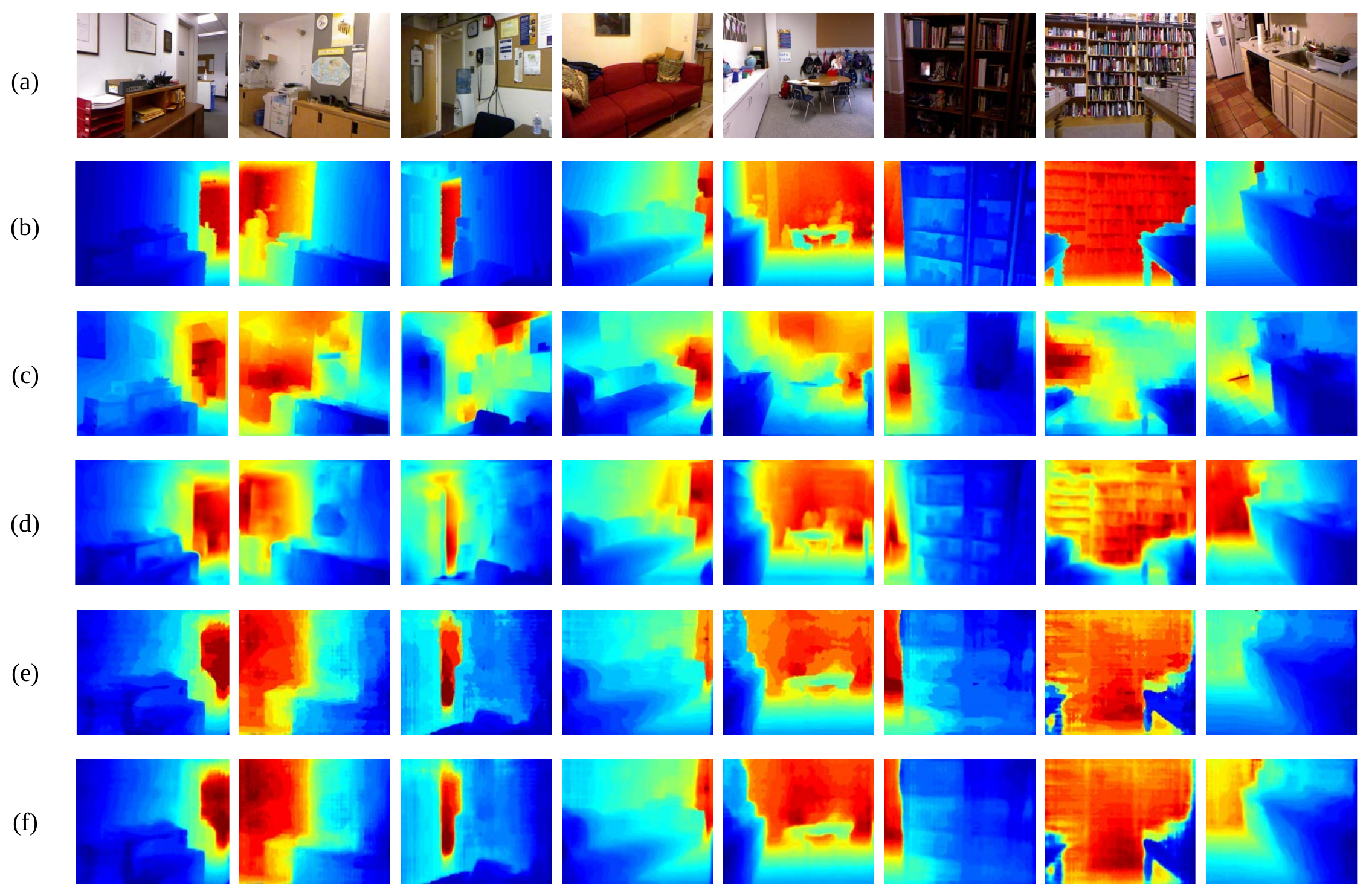}
	\end{center}
	\caption{Some depth estimation results on the  NYUD2 dataset.
    (a) RGB Input; (b) Ground-truth depth; (c) Results of Liu et al. \cite{LiuSLR15}; (d) Results of Eigen et al. \cite{Eigen15};
(e) Results of our model without fully-connected CRFs; (f) Results of our model with fully-connected CRFs.}
	\label{fig:visual}
\end{figure*}

We train our model using the entire raw training data specified in the official train/test distribution and test on the standard 654 test images. We discretize the depth values into 100 bins in the log space. We set the parameter $\alpha$ of the information gain matrix to be 0.2. The fully connected CRFs are applied as post-processing. The results are reported in Table \ref{table:state-of-art}. The first row is the result in \cite{Wang_2015_CVPR} which jointly performs depth estimation and semantic segmentation. The second row is the result of deep convolutional neural fields (DCNF) with fully convolutional network and super-pixel pooling in \cite{LiuSLR15}. The third row is the result of nerual regression forest (NRF) in \cite{RoyT16}. The fourth row is the result in \cite{Eigen15} which performs depth estimation in a multi-scale network architecture. The fifth row is the result in \cite{laina2016deeper} which applies an upsampling scheme. The last row is depth estimation result by our model. As we can see from the table, our deep fully convolutional residual network with depth label classification achieves state-of-the-art performance of 4 evaluation metrics. We also show some qualitative results in Fig. \ref{fig:visual}, from which we can see our method yields better visualizations in general.

\begin{table}
	\renewcommand\arraystretch{0.65}
	\caption{Test results on the SUN RGB-D dataset for cross-dataset evaluation. The first 2 rows are results by recent depth estimation models. The last row is the result of our approach.}
	\centering
	\label{table:cross_dataset}
	\begin{tabular}{@{\hskip 0.05cm}c@{\hskip 0.07cm}c@{\hskip 0.07cm}c@{\hskip 0.07cm}c @{\hskip 0.26cm}c@{\hskip 0.15cm}c@{\hskip 0.15cm}c}
        \noalign{\smallskip}
		\hline
		\noalign{\smallskip}
						& \multicolumn{3}{c}{\small Accuracy}    & \multicolumn{3}{c}{\small Error}  \\
		                 & \small $\delta<1.25$   &\small $\delta<1.25^2$    &\small $\delta<1.25^3$    &\small rel    &\small log10  &\small rms  \\
		\noalign{\smallskip}

		\hline
		\noalign{\smallskip}
   \small Liu \cite{LiuSLR15}     &\small  35.6\%   &\small 57.6\%   &\small 83.1\%   &\small 0.316    &\small  0.161  &\small  0.931 \\
		\noalign{\smallskip}
   \small Laina \cite{laina2016deeper} &\small 53.9\%   &\small  70.3\%   &\small \bf 89.0\%   &\small  0.279  &\small  0.138  &\small  0.851 \\
   		\noalign{\smallskip}
   \small Ours     &\small \bf 56.3\%   &\small \bf 72.7\%   &\small  88.2\%   &\small \bf 0.256    &\small \bf 0.127  &\small \bf 0.839 \\
		\noalign{\smallskip}
		\hline
		
	\end{tabular}
\end{table}

\subsubsection{KITTI}
\begin{figure*}
	\begin{center}
		\includegraphics[scale=.45]{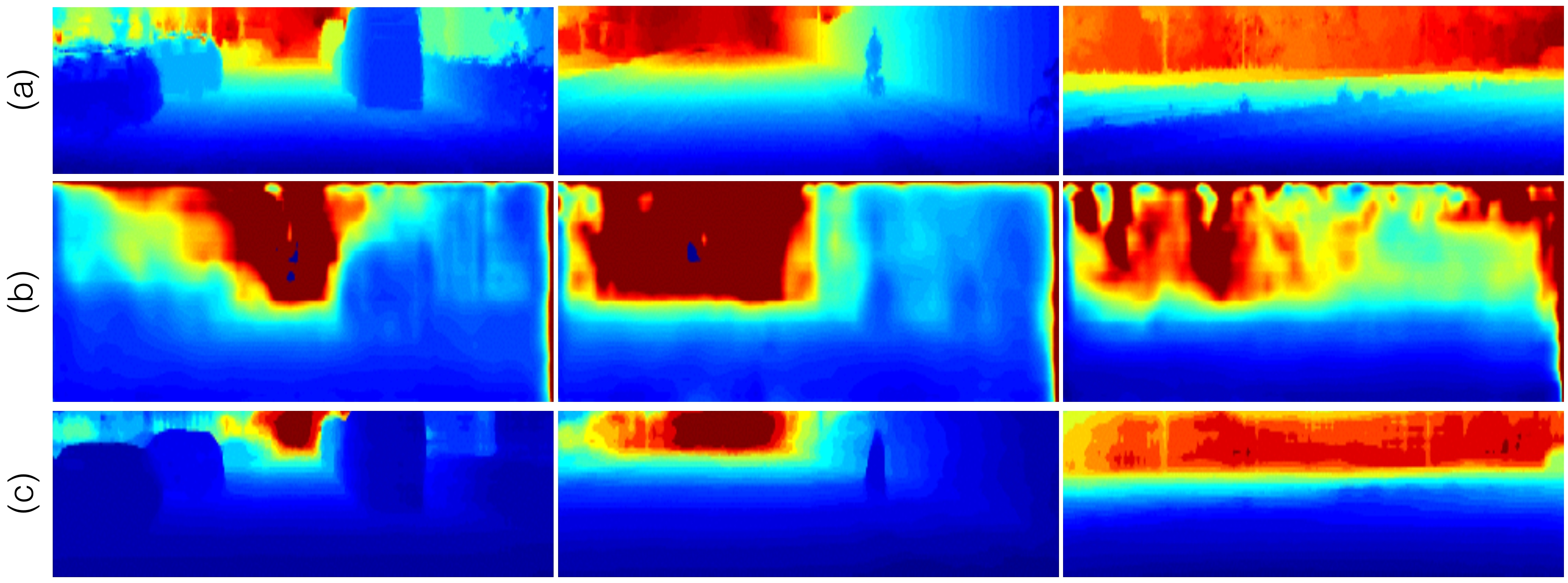}
	\end{center}
	\caption{Some depth estimation results of the KITTI dataset. The first row are the ground-truth depths, the second row are the results by \cite{garg2016unsupervised}, the last row are the results by our approach.}
	\label{fig:kitti_visual}
\end{figure*}

We train our model on the same training set in \cite{godard2016unsupervised} which contains 33131 images and test on the same 697 images in \cite{EigenPF14}. But different from the depth estimation method proposed in \cite{godard2016unsupervised} which applies both the left and right images in stereo pairs, we only use the left images. The missing values in the ground-truth depth maps are ignored during both training and evaluation. The depth values are discretized into 50 bins in the log space. We set the parameter $\alpha$ of the information gain matrix to be 0.5 and apply fully connected CRFs as post-processing. In order to compare with the recent state-of-the-art results, we cap the maximum depth into both 80 meters and 50 meters and present the results in Table~\ref{table:stat-of-art_kitti}. We can see from Table~\ref{table:stat-of-art_kitti} that our method outperforms the rest methods significantly. Some qualitative results are illustrated in Fig. \ref{fig:kitti_visual}. Our approach yields visually better results.

\subsubsection{Cross-dataset evaluation}
In order to show the generalization of our proposed method, we train our model on the raw NYUD2 dataset and test on the SUN RGB-D dataset \cite{Song_2015_CVPR}. The SUN RGB-D is an indoor dataset contains 10335 RGB-D images captured by four different sensors. We only select 500 images randomly from the test set for cross-dataset evaluation. The SUN RGB-D contains 1449 images from the NYUD2 dataset. Our selected testset exlucdes all the images from the NYUD2. We compare our method with Liu et al. \cite{LiuSLR15} and Laina et al. \cite{laina2016deeper}. We use the trained models and evaluation codes released by these authors. The results are illustrated in Table~\ref{table:cross_dataset}. We can see that our method can reach satisfactory results on different dataset, and outperforms other methods.

\section{Conclusion}\label{sec:con}
We have presented a deep fully convolutional residual network architecture for depth estimation from  single monocular images.
We have made use of the recent deep residual networks, discretized continuous depth values into different bins and formulated depth estimation as a discrete classification problem. By this formulation we can easily obtain the confidence of a prediction which can be applied during training via information gain matrices as well as post-processing via fully-connected CRFs. We have shown that our discretization approach surprisingly performs well.

Note that the proposed network can be further improved by applying the techniques that have been previously explored. For example, it is expected that
\begin{itemize}
    \item
Multi-scale inputs as in \cite{Eigen15} would improve our result.
\item  Concatenating the mid-layers' outputs may better use the low-, mid-layers information as in \cite{BharathCVPR2015}.
\item  Upsampling the prediction maps as in \cite{long_shelhamer_fcn} would be beneficial too.
\end{itemize}
We  leave these directions in our future work.

\bibliographystyle{IEEEtran}
\newpage
\end{document}